\documentclass[12pt,twoside]{article}   %For printing on two sides of page
\usepackage[super,sort&compress,comma]{natbib}
\usepackage{fancyhdr}		%Gives headers and footers defined below
\usepackage[section]{placeins}   %
\usepackage{graphicx}
\usepackage[para]{threeparttable} % table
\usepackage{booktabs} %toprule
\usepackage{xcolor}
\newcommand{\cmark}{\ding{51}}
\newcommand{\xmark}{\ding{55}}
\usepackage{pifont}
\makeatletter \renewcommand\@biblabel[1]{$^{#1}$} \makeatother
\newcommand*\xbar[1]{%
   \hbox{%
     \vbox{%
       \hrule height 0.2pt % The actual bar
       \kern0.5ex%         % Distance between bar and symbol
       \hbox{%
         \kern-0.1em%      % Shortening on the left side
         \ensuremath{#1}%
         \kern-0.1em%      % Shortening on the right side
       }%
     }%
   }%
} 

\newcommand{\cen}[1]{\begin{center} #1 \end{center}}

\usepackage[mathlines]{lineno}
\usepackage{amssymb}
\usepackage{hyperref}
\hypersetup{ colorlinks,
    citecolor=blue,
    filecolor=blue,
    linkcolor=blue,
    urlcolor=blue
}

\usepackage{multirow}
\usepackage[T1]{fontenc}
\usepackage[font=small,labelfont=bf,tableposition=top]{caption}

\DeclareCaptionLabelFormat{andtable}{#1~#2  \&  \tablename~\thetable}

\usepackage{xcolor}
\definecolor{gray}{rgb}{0.6,0.6,0.6}
\definecolor{red}{rgb}{0.85,0,0}
\definecolor{green}{rgb}{0,0.85,0}
\definecolor{blue}{rgb}{0,0,0.85}
\definecolor{beige}{rgb}{0.92,0.87,0.78}
\usepackage[all]{hypcap}    %causes link to figures to go to figure, not caption
\newcommand{\X}{\mathbf{x}}
\newcommand{\N}{\mathbf{N}}

\newcommand{\n}{\mathbf{n}}
\def \<= {%
  \leq}%
\def \>= {%
  \geq}%
\newcommand{\etal}{\textit{et al}.~}

\usepackage{mathtools,xparse}
\usepackage{amsmath,bm}
\NewDocumentCommand{\normL}{ s O{} m }{%
  \IfBooleanTF{#1}{\norm*{#3}}{\norm[#2]{#3}}_{L_2(\Omega)}%
}

\usepackage{multicol}
\begin{document}

\cen{\sf {\Large {\bfseries FDRN: A Fast Deformable Registration Network \\
for Medical Images }} \\  
\vspace*{8mm}
Kaicong Sun$^*$, Sven Simon \\
\vspace*{2mm}
Institute of Parallel and Distributed Systems, University of Stuttgart, %Universit\"atsstra\ss e 38, Stuttgart,
 Stuttgart, Germany
%\vspace{5mm}\\
%Version typeset \today\\
}

\pagenumbering{arabic}
\setcounter{page}{1}
%\fancypagestyle{plain}
%\pagestyle{plain}
\pagestyle{fancy}
\fancyfoot{}
% Set the right side of the footer to be the page number
%\fancyfoot[R]{\thepage}
\cfoot{\thepage}
$^*$Corresponding author. Email: kaicong.sun@ipvs.uni-stuttgart.de
\\
% note, probably best not to use a student's e-mail as it won't be valid for
% very long.

\begin{abstract}
\noindent {\bf Purpose:} Deformable image registration is a fundamental task in medical imaging. Due to the large computational complexity of deformable registration of volumetric images, conventional iterative methods usually face the tradeoff between the registration accuracy and the computation time in practice. In order to boost the performance of deformable registration in both accuracy and runtime, we propose a fast unsupervised convolutional neural network for deformable image registration.\\
{\bf Methods:} The proposed registration model FDRN possesses a compact encoder-decoder network architecture which employs a pair of fixed and moving images as input and outputs a three dimensional displacement vector field (DVF) describing the offsets between the corresponding voxels in the fixed and moving images. In order to efficiently utilize the memory resources and enlarge the model capacity, we adopt additive forwarding instead of channel concatenation and deepen the network in each encoder and decoder stage. To facilitate the learning efficiency, we leverage skip connection within the encoder and decoder stages to enable residual learning and employ an auxiliary loss at the bottom layer with lowest resolution to involve deep supervision. Particularly, the low-resolution auxiliary loss is weighted by an exponentially decayed parameter during the training phase. In conjunction with the main loss in high-resolution grid, a coarse-to-fine learning strategy is achieved. Last but not least, we involve a proposed multi-label segmentation loss to improve the network performance in Dice score in case the segmentation prior is available. Comparing to the segmentation loss using average Dice score, the proposed segmentation loss does not require additional memory in the training phase and improves the registration accuracy efficiently. \\
{\bf Results:} We evaluated FDRN on multiple brain MRI datasets from different aspects including registration accuracy, model generalizability, and model analysis. Experimental results demonstrate that FDRN performs better than the state-of-the-art registration method VoxelMorph by 1.46\% in Dice score in LPBA40. In addition to LPBA40, FDRN obtains the best Dice and NCC among all the investigated methods in the unseen MRI datasets including CUMC12, MGH10, ABIDE, and ADNI by a large margin.\\
{\bf Conclusions:} The proposed FDRN provides better performance than the existing state-of-the-art registration methods for brain MR images by resorting to the compact autoencoder structure and efficient learning. Additionally, FDRN is a generalized framework for image registration which is not confined to a particular type of medical images or anatomy. 

\end{abstract}

\iffalse
The table of contents is for drafting and refereeing purposes only. Note
that all links to references, tables and figures can be clicked on and
returned to calling point using cmd[ on a Mac using Preview or some
equivalent on PCs (see View - go to on whatever reader).
\tableofcontents

\newpage

\setlength{\baselineskip}{0.7cm}      %double spacing		

\setcounter{page}{1}

\fi

\section{INTRODUCTION}
Deformable medical image registration is an approach to establish dense spatial correspondence between a pair of digital images based on the local morphological structures. Deformable registration is widely applied to many medical applications such as detecting temporal anatomical changes of individuals, analyzing variability across populations, and multi-modality fusion. In recent decades, several advancements in deformable image registration have been made~\cite{survey2,survey3,survey4,survey5}. Most of the existing conventional algorithms optimize an objective function which consists of a data term $D$ and a regularization term $R$ formulated as following:
\begin{linenomath}
\begin{equation}\label{Objective}
J = D(F(\X), \phi\circ M(\X)) + \lambda R(\phi).
\end{equation}
\end{linenomath}
The data term $D$ measures the alignment between the fixed image $F$ and the transformed moving image $\phi\circ M$ with $\phi$ being the dense deformation field which maps the coordinates of $F$ to the coordinates of $M$ and $\circ$ being the resampling operation. $\X$ denotes the 3D spatial coordinate in domain $\{\Omega~|~\Omega\subset \mathbb{R}^3\}$. The most commonly used data terms are based on, e.g., $L2$ error norm~\cite{Beg2}, mutual information~\cite{MI}, and cross-correlation~\cite{Avants}. As deformable registration is a highly ill-posed problem, regularization $R$ is inevitably required to involve the prior knowledge of the transformation such as smoothness into the objective function. The weighting parameter $\lambda$ balances the tradeoff between the data term $D$ and the regularization term $R$. In general, the deformation field $\phi$ is modeled either by the displacement vector field d(x) or the velocity vector field $v(\X, t)$. The former category models the spatial transformation as a linear combination of the identity transform $\X$ and the DVF: $\phi = \X+d(\X)$ with $d(\X)$ being the displacement vector field (DVF) which represents the spatial offsets between the corresponding voxels in the fixed and the moving images. Particularly, \cite{Bajcsy, Gee, Davatzikos} estimate the deformation based on the linear elastic models. In~\cite{Rueckert, Kybic, Sdika}, the deformation field is described by cubic B-spline. Thirion proposes Demons~\cite{Thirion} by introducing diffusion model in image registration. Generally, the DVF-based deformation model can not guarantee an inverse consistency such that interchanging the order of the two input images, the obtained transformation may not match the inverse of the counterpart. In contrast to the displacement-based vector field, the latter one concerns the invertibility of the transformation. Specially, deformable registration is considered as a variational problem and $\phi$ is formulated as an integral of a velocity vector
field $v(\X, t)$. Many variants have been proposed~\cite{Beg1, Beg2, Avants, Ashburner2} imposing biomedical constraints such as diffeomorphism, topology preservation, inverse consistency, and symmetry on the deformation field. In~\cite{Beg2}. Beg~\etal present the large deformation diffeomorphic metric mapping (LDDMM) to solve a global variational problem in the space of smooth velocity vector field. Avants~\etal\cite{Avants} propose the symmetric normalization method (SyN) using cross-correlation as the similarity measure. However, due to the huge computational demand for volumetric medical images, tackling practical problems by conventional methods could be extremely slow. 

In contrast to the traditional optimization-based methods which adopt iterative updating scheme, learning-based methods are usually trained offline based on a large-scale dataset. As long as the models are well-trained, predicting the transformation between unseen images performs solely forward propagation and consumes significantly less computation time. In other words, the learning-based methods transfer the burdens of computation to offline training and the well-trained model is dedicated to the specific application learned from the training dataset. Abundant learning-based studies on medical image registration have been conducted in the last two decades~\cite{rueckert1, rueckert2, krebs, gutierrez1, gutierrez2, kim, wang}. In particular, Kim~\etal\cite{kim} present a patch-based deformation model using sparse representation. In~\cite{gutierrez2}, Guti\'errez-Becker~\etal formulate the prediction of the transformation parameters as supervised regression using the gradient boosted trees. More recently, deep learning has attracted increasing attention in the field of medical image registration due to the prominent capability of feature extraction~\cite{survey5}. Based on the supervision type, the existing deep learning models can be categorized into supervised, unsupervised/self-supervised, weakly supervised, deeply supervised, and dual supervised. Specially, \cite{fan, sokooti, yang, rohe} adopt deep convolutional neural network (CNN) in a supervised manner which require the ground truth of the deformation field during the training phase. However, supervised models usually suffer from the inaccuracy of the ground truth deformation field in the training datasets. In contrast to supervised learning, weak supervision employs higher-level correspondence information such as corresponding anatomical structural masks or landmark pairs which are more practical to obtain~\cite{hu2018}. Some works~\cite{deVos, Balakrishnan, Li} propose CNN models based on unsupervised learning by optimizing the similarity match between the fixed image and the transformed moving image by resorting to the spatial transformer network (STN)~\cite{Jaderberg}. Specially, Balakrishnan~\etal\cite{Balakrishnan} introduce a CNN framework based on the UNet structure~\cite{unet} adopting local cross-correlation as the data term. In~\cite{Li}, Li~\etal construct a multi-resolution registration model which contains three losses constraining the DVF at different spatial resolutions to maximize the similarity at different resolutions. In~\cite{V2}, the authors extend their previous work~\cite{Balakrishnan} by employing average Dice score of the segmented regions as the auxiliary loss. Fan~\etal\cite{fan} present a modified UNet named BIRNet which uses multi-channel input and hierarchical loss based on dual supervision. Particularly, the loss consists of a supervised part which measures the deviation of the deformation field and an unsupervised part which guides the similarity match between the fixed and the warped moving image. However, due to the computational complexity, BIRNet requires 17.4s to register a pair of images of size $220\times 220\times 184$. 

Autoencoder network has achieved promising performance in multiple medical image processing applications such as lesion detection, tumor segmentation, and image denoising. By propagating the high-frequency information from the encoder path directly to the decoder path, autoencoder architecture is able to preserve local information and meanwhile obtain large receptive field. In this paper, we propose a fast deformable registration network FDRN based on the autoencoder backbone. One challenge of 3D image registration is the huge memory consumption which prohibits the deepening of the network. In order to address the issue, instead of using channel concatenation as VoxelMorph~\cite{V2}, we adopt additive forwarding  between the encoder and the decoder layer and the saved training memory is exploited to deepen the network. In order to enhance the learning efficiency of our deep model, we leverage deep supervision at the bottom layer with smallest resolution to guide the convergence of the network and adopt skip connection in the encode and decoder stages to enable residual learning. Aiming for utilizing the available segmentation prior, we propose a multi-label segmentation loss which improves the registration accuracy efficiently without inducing additional memory cost in the training. In the experiment, we demonstrate that the proposed FDRN achieves better performance than the investigated state-of-the-art registration methods for brain MR images. Although in this paper FDRN is evaluated on the brain MRI datasets, by resorting to the general formulation of the loss function, FDRN is able to perform deformable registration on other anatomic structures or images types. 

The paper is organized as follows. In Section~\ref{sec:methods}, we introduce the proposed FDRN model. Section~\ref{sec:results} demonstrates the evaluation of FDRN from different aspects including registration performance and model analysis. 
In Section~\ref{sec:disconc}, we discuss the insights of the experiments and conclude our work.
  
\section{MATERIALS AND METHODS}\label{sec:methods}
\begin{figure*}
	\centering
	 \includegraphics[scale=0.75]{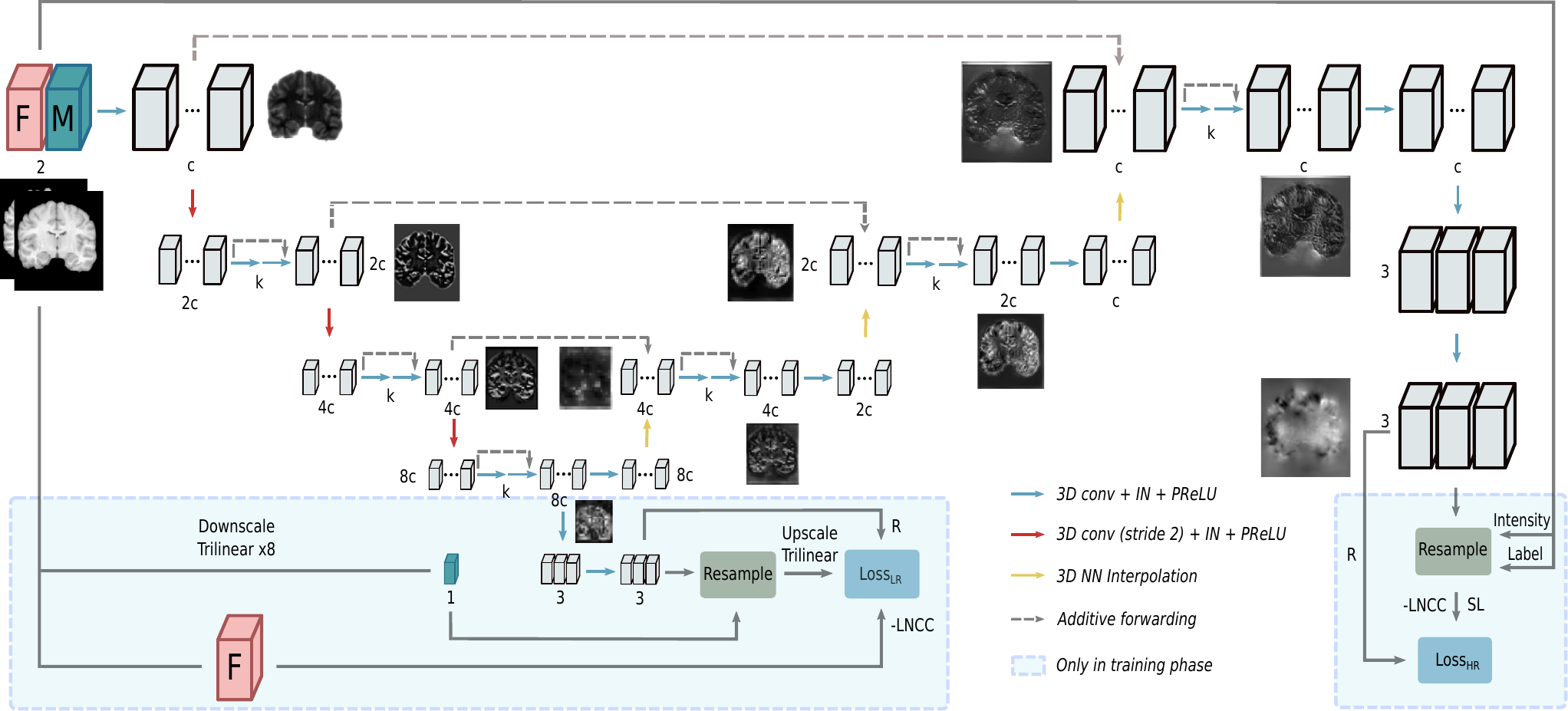}		
	\caption{Schematic illustration of the structure of the proposed FDRN. One feature map is shown adjacent to the associated layer for demonstration purpose. Variable $c$ depicts the amount of channels in the first layer and $k$ represents the number of convolutions at each encoder-decoder stage. For our baseline model, we choose $c=8, k=1$ and we set $c=16, k=2$ in FDRN.}
	\label{fig:FDRN}
	\vspace{2em}
\end{figure*} 
The proposed FDRN is based on a compact encoder-decoder structure as demonstrated in Fig.~\ref{fig:FDRN}. FDRN has a two-channel input which consists of a fixed and moving image pair and outputs a three-channel displacement vector field (DVF) which describes the offsets between the corresponding voxels in the fixed and moving images. Based on the output DVF $d(\X)$, the moving image $M$ is resampled at the transformed nonvoxel location $\phi(\X) = \X + d(\X)$ and a similarity measure between the fixed image $F$ and the transformed moving image $\phi\circ M$ is optimized. Hence, the registration network can be trained in an unsupervised scheme regardless of the ground-truth DVF. Particularly, the encoder path extracts the features at different resolutions and meanwhile enlarges the receptive field by convolutions with stride 2. Each convolution is followed by the instance normalization and PreLU. Skip connection is utilized at each encoder and decoder stage to enable residual learning and prevent from gradient vanishing. Due to the fact that the registration of low-resolution (LR) images is easier to learn, we involve an auxiliary loss at the bottom of the encoder path to punish the misalignment of the large structures in the LR image pairs. Specially, the weight of the LR loss is gradually decayed along with the training and the model is fine tuned fully based on the main loss in high-resolution (HR) grid such that the LR auxiliary loss guides the convergence of the network in the early training phase which imitates the coarse-to-fine registration strategy applied in the conventional multi-resolution registration methods. On the decoder side, the extracted features are fused and the feature maps are enlarged to restore the original image dimension. Particularly, instead of using deconvolution, we combine nearest neighbor (NN) interpolation with 3D convolution to circumvent the checkerboard artifacts induced by deconvolution. In order to save memory and preserve the high-frequency features, we perform additive forwarding from the encoder path to the corresponding decoder path. Last but not least, we introduce a multi-label segmentation loss (SL) to further boost the performance of registration. Comparing to the average Dice score employed in~\cite{V2}, the proposed SL does not require additional memory in the training regardless of the number of classes.

For the sake of clarity, we indicate different structure variants of our model by $c~\text{-}~k$ in the latter formulation where $c$ represents the amount of channels in the first layer and $k$ denotes the number of convolutions at each encoder and decoder stage. Combining the loss functions under different resolutions, we yield the overall loss formulated as 
\begin{linenomath}
\begin{equation}\label{Joint}
L_{overall} = (1-\lambda) L_{HR} + \lambda L_{LR}.
\end{equation}
\end{linenomath}
The parameter $\lambda$ is exponentially decreased from 0.5 to 0 such that the HR loss dominates the learning gradually in the training. 
The $L_{HR}$ and $L_{LR}$ are defined as
\begin{linenomath}
\begin{equation}\label{Loss}
\begin{split}
L_{HR} &= -D(F_{HR}(\X), \phi_{HR}\circ M_{HR}(\X)) + \alpha_1 R(\phi_{HR}) + \alpha_2 SL(\phi_{HR}),\\
L_{LR} &= -D(F_{HR}(\X), [\phi_{LR}\circ M_{LR}(\X)]_{\uparrow}) + \alpha_3 R(\phi_{LR}),
\end{split}
\end{equation}
\end{linenomath}
where $D$ denotes the data fidelity term, $\phi_{HR}$ and $\phi_{LR}$ indicate respectively the deformation field in the HR and LR grid. Inspired by~\cite{V2}, we adopt local normalized cross-correlation (LNCC) as expressed in Eq.~\eqref{dissim} to quantify the similarity measure. Comparing to normalized cross correlation (NCC), it turns out that LNCC converges faster and better for large training patchsize. $[\,\cdot\,]_{\uparrow}$ represents the upscaling operation and in this work, we use trilinear interpolation.
$\alpha_1, \alpha_3$ are the weights of the regularization terms and $\alpha_2$ indicates the weighting parameter of the segmentation loss.
\begin{linenomath}
\begin{equation}\label{dissim}
D(F,\phi\circ M)= \sum_{\X_i}^{\Omega}\frac{\big(\sum_{\X}^{\Omega_i}(F(\X)-\xbar{F}_{\Omega_i})(\phi\circ M(\X)-\xbar{\phi\circ M}_{\Omega_i})\big)^2}{(\sum_{\X}^{\Omega_i}(F(\X)-\xbar{F}_{\Omega_i})^2)(\sum_{\X}^{\Omega_i}(\phi\circ M(\X)-\xbar{\phi\circ M}_{\Omega_i})^2)}
\end{equation}
\end{linenomath}
In the formulation of LNCC in Eq.~\eqref{dissim}, $\xbar{F}_{\Omega_i}$ denotes the mean of the local region $\Omega_i$ centered at voxel $\X_i$ with size of $n^3$. $\xbar{\phi\circ M}_{\Omega_i}$ represents the mean of the corresponding region in the transformed moving image. In this work, we choose $n=9$.

The regularization $R$ imposes smoothness constraint on the deformation field $\phi$. As $\phi$ is a linear combination of the identity transform $\X$ and the expected DVF, we apply the constraint directly on the DVF $d(\X)$ by
\begin{linenomath}
\begin{equation}\label{tv}
R(d(\X))=\sum\limits_{S_k}||d(\X)-S_k d(\X)||^2_2,
\end{equation}
\end{linenomath}
where $S_k$ indicates the shifting operator along $(u,v,w)$ direction by vector $k$ with $k = \{(u,v,w)~|~u,v,w\in\{0,1\}\}$ and $||\cdot||^2_2$ represents the L2-norm. Comparing to the L1-norm total variation (TV), L2-norm is differentiable at 0 and leads to a considerable faster convergence. 

The segmentation loss SL serves to punish the misalignment between the labels of the fixed image $L_F$ and the transformed moving image $\phi\circ L_M$. In~\cite{V2}, average Dice score (ADS) is adopted to regularize the DVF based on the segmentation labels. In fact, multi-label dice score is originally used for segmentation applications. Acting as the regularization term for volumetric deformable registration, ADS has two major drawbacks. First, the leverage of ADS in the loss function for large 3D image induces noticeable additional memory cost during the training which prohibits the deepening of the network. Second, the memory consumed by ADS increases linearly with the segmentation classes. For instance, a label volume of size $160\times208\times176$ with $56$ classes as brain MRI dataset LPBA40 requires 1.3GB during training. 
In order to tackle this issue, we propose a multi-label SL as
\begin{linenomath}
\begin{equation}\label{R2}
SL(\phi)=\frac{(c_1+1)|L_{F}-\phi\circ L_{M}|_1}{|L_F|_1+|\phi\circ L_M|_1+c_1|L_{F}-\phi\circ L_{M}|_1+c_2},
\end{equation}
\end{linenomath}
where $L_F$ and $L_M$ represent respectively the labels of the fixed image and the moving image. $c_1$ and $c_2$ are nonnegative constants. Specially, $c_1$ weights the punishment of the inconsistency between $L_F$ and $\phi\circ L_{M}$ and $c_2$ serves to prevent zero division. Comparing to ADS, the proposed segmentation loss does not require extra memory but depends on the value of the individual label.

For the transformation of the moving image $\phi \circ M(\X)$, we adopt the trilinear interpolation of $M(\X)$ at the nonvoxel locations $\phi$. Specially, the resampling of $M(\X)$ at the transformed location $\phi$ is formulated as
\begin{linenomath}
\begin{equation}\label{spatial transform}
\phi\circ M(\X) = \sum\limits_{\mathclap{\n\in \N(\phi(\X))}} M(\n)\prod\limits_{\mathclap{m\in\{u,v,w\}}}(1-|\phi_m(\X)-\n_m|), 
\end{equation}
\end{linenomath}
where $\N(\phi(\X))$ denotes the coordinates of the neighbors of the transformed nonvoxel location in the moving image $M(\X)$ and $\n$ represents the coordinates of the individual neighbor. $m$ is an indicator and iterates over the dimensions of the moving image.

\section{RESULTS}\label{sec:results}
In this section, we conduct experiments to evaluate the proposed FDRN from different aspects. Particularly, in Section~\ref{LPBA40} we train and evaluate FDRN on the LONI LPBA40 dataset~\cite{LPBA40} which contains T1-weighted brain MR images. We compare FDRN with the state-of-the-art registration methods including the traditional method symmetric image normalization SyN~\cite{Avants}, the deep learning-based methods Li~\etal\cite{Li} and VoxelMorph~\cite{Balakrishnan}. In Section~\ref{AA}, we validate the well-trained FDRN on other unseen MRI datasets including CUMC12~\cite{klein}, MGH10~\cite{klein}, ABIDE~\cite{ABIDE}, and ADNI~\cite{ADNI}. Section~\ref{sec:MA} consists of the model analysis.

\subsection{Datasets and Preprocessing}\label{sec:Dataset}
The LPBA40 dataset contains brain MR images of 40 neurologically intact nonepileptic subjects with segmentation labels for 56 brain regions including the hippocampi. All of these MR images were firstly registered to the Montreal Neurological Institute (MNI) space using affine transformation based on the ICBM152 template~\cite{unbiased} as preprocessing. The registered images were then cropped to the size of $160\times208\times176$. We partitioned the 40 cropped images into 30, 4, and 6 for training, validation, and testing, respectively. In addition, we conducted experiments on the unseen CUMC12, MGH10, ABIDE and ADNI MRI datasets to evaluate the generalizability of the well-trained FDRN. Specially, CUMC12 contains 12 MR images with 128 labeled regions and MGH10 consists of 10 subjects with 74 segmented regions. We randomly selected 10 images individually from ABIDE and ADNI which do not contain segmentation labels. The experimental results are demonstrated in Section~\ref{LPBA40} and Section~\ref{AA}.

\subsection{Evaluation Metrics}\label{Metrics}
In addition to NCC which measures the similarity in cross correlation, Dice score~\cite{Dice} is adopted to quantify the overlap of labels for each segmented region by
\begin{linenomath}
\begin{equation}\label{dice}
Dice(A, B)=\frac{2|A\cap B|_1}{|A|_1+|B|_1},
\end{equation}
\end{linenomath} 
where $A$ and $B$ indicate binary images which represent the individual label in the fixed and moving image, respectively. Furthermore, we leverage the perception-based metric, structural similarity index measure (SSIM)~\cite{ssim}, to aid the assessment in visual perception in terms of luminance, contrast and structure:
\begin{linenomath}
\begin{equation}\label{ssim}
SSIM(A, B) = \frac{(2\mu_A\mu_B+c_1)(2\sigma_{AB}+c_2)}{(\mu_A^2+\mu_B^2+c_1)(\sigma_A^2+\sigma_B^2+c_2)},
\end{equation}
\end{linenomath} 
where $\mu_A, \mu_B, \sigma_A, \sigma_B$ denote the mean and standard deviation of the image $A$ and $B$. $\sigma_{AB}$ indicates the covariance between image $A$ and $B$. $c_1, c_2$ are small positive constants to stabilize the division with weak denominator. An SSIM of 1 indicates a perfect anatomical match.

\subsection{Implementation Details}\label{sec:Implementation}
The proposed FDRN was implemented with a Pytorch backend. As FDRN was trained in an unsupervised manner, in order to ensure the existence of the corresponding voxels in the input pair especially at the image boarders during training, we fed the network with the whole image of size $160\times208\times176$. Due to the memory limitation, a mini-batch of size 1 was used and hence, there were 870 iterations in each epoch. We employed Adam with $\beta_1 = 0.9, \beta_2 = 0.999$ as the optimizer. The initial learning rate was set as 0.002 and multiplied by 0.9 every 1000 iterations until decreased to 0.0001 over 70 epochs. The weighting parameters $\alpha_1, \alpha_3$ for the regularization $R$ were set as $\alpha_1 = 1\times10^{-8}, \alpha_3 = 8\alpha_1$. The weight $\lambda$ of $L_{LR}$ was implemented as $0.5^{(1+i/1000)}$ with $i$ being the index of the iteration and $0.5$ as the initial weight. The weight of the segmentation loss SL was tuned as $\alpha_2=0.2$ and the parameters in SL were set as $c_1=10, c_2=10^{-9}$. A detailed analysis of hyperparameters $\alpha_2$ and $c_1$ is carried out in Section~\ref{sec:MA}. It is worthy noting that the above-mentioned hyperparameters were tuned on the validation dataset in LPBA40 and the ones generating the best Dice score were selected. The experiments were performed on the NVIDIA GeForce GTX 1080 Ti with 11GB GDDR5X and the Intel(R) Xeon(R) E5-2650 v2 CPU.

\begin{figure*}
%\vspace*{-0.7cm}
	\centering
	  \hspace*{-0.4cm}\includegraphics[scale=0.88]{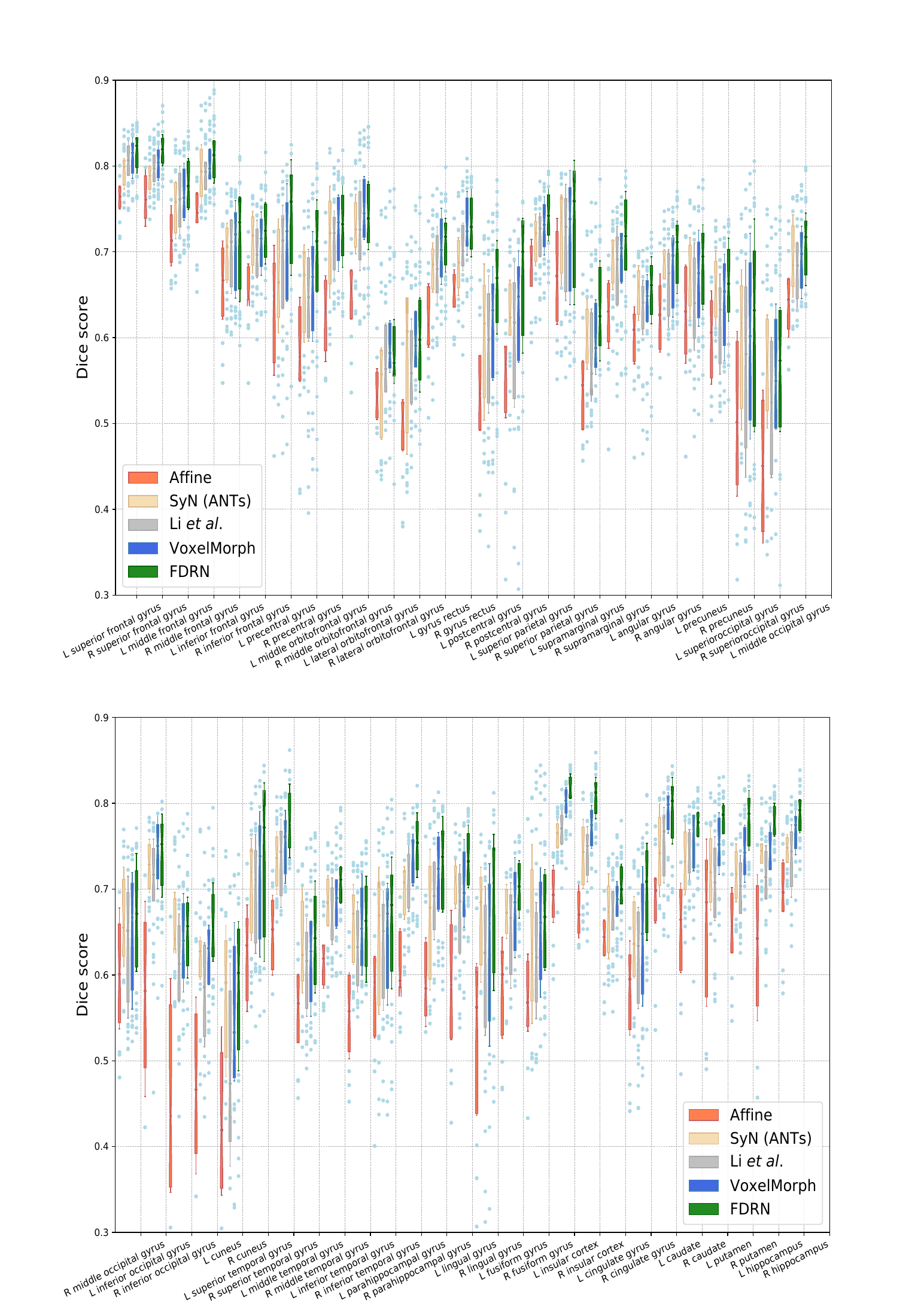}		
	\caption{Boxplots of the average Dice scores of 54 labeled anatomical regions for the 30 testing image pairs from the publicly available LPBA40 brain MRI dataset. Affine represents the results of the globally registered images based on affine transformation. SyN indicates the results by the traditional symmetric image normalization method which was implemented by ANTsPy. Li~\etal and VoxelMorph are the learning-based state-of-the-art registration methods. It is shown that the proposed FDRN performs best in Dice score for nearly all the regions.}
	\label{fig:Boxplot}
\end{figure*}

\subsection{Evaluation on LPBA40 Dataset}\label{LPBA40}
We conducted comparison with the existing state-of-the-art deformable registration methods SyN~\cite{Avants}, Li~\etal\cite{Li} and VoxelMorph~\cite{V2} on the public LPBA40 dataset. We reproduced and trained Li's model, VoxelMorph in Pytorch according to their original paper. Specially, for Li's method, we used the full image as the network input with a mini-batch of size 1 and tuned the weight of the total variation regularization as $\lambda=1\times10^{-9}$ for best Dice performance. With regard to VoxelMorph, we utilized NLCC as the main loss, L2 smoothness as the regularization and average Dice score as the auxiliary loss. The weight of the Dice loss was tuned as 0.1 and the model was trained over 70 epochs. We employed ANTsPy, the Python wrapper for the Advanced Normalization Tools (ANTs)~\cite{ANTs}, to implement SyN. Particularly, we adopted cross correlation (CC) as the similarity measure and instead of using the default iterations $(40, 20, 0)$, we set the iterations as $(100,40,10)$. To achieve better accuracy, we set the sampling bins as 60 instead of the default 32. 30 MR images were used as the training data and the remaining 10 images were utilized for validation and testing. Every permutation of pairs out of the 30 images (total of 870 permutations) was selected as the input during the training phase. In the testing phase, each of the 6 images was chosen as the fixed image and the rest 5 images were registered to it (total of 30 pairs of images). We quantified the performance of SyN, Li's method, VoxelMorph, our baseline model, and FDRN using Dice score, NCC, and runtime as summarized in Table~\ref{tab:LPBA40}. Particularly, we compared with VoxelMorph with and without the ADS loss. As depicted, all the CNN-based methods perform hundreds times faster than the traditional SyN. Comparing to VoxelMorph, our baseline model obtains comparable Dice and nearly halves the inference time by discarding the channel concatenation. The proposed FDRN improves the Dice and NCC efficiently by enlarging the network capacity. Comparing to VoxelMorph with ADS, our 16-2 consumes similar training memory (about 10.8GB) and inference time and achieves a performance gain of 1.46\% in Dice score. It is worth noting that the runtime is accumulated purely for the registration step without concerning the preprocessing and image loading. Additionally, we present the Dice score of each anatomical structure labeled in LPBA40 obtained by different methods in the boxplots shown in Fig.~\ref{fig:Boxplot}. We can observe that FDRN performs best among the investigated methods for nearly all the labeled regions.

\begin{table*}
	\centering
\begin{threeparttable}[b]
	\setlength{\tabcolsep}{1pt}
	\caption{Comparison of different registration methods on the testing images with size of 160$\times$208$\times$176 in the LPBA40 MRI dataset by average Dice score, NCC, and runtime. ADS: Average Dice score; SL: Segmentation loss. Best results are in bold.}
	\label{tab:LPBA40}
	\begin{tabular}{c | c | c | c  | c  c  | c  c | c  c}
		\toprule
		\textbf{} & \textbf{Affine} & \textbf{SyN~\cite{Avants}} & \textbf{Li \textit{et al.}~\cite{Li}} &  \multicolumn{2}{c}{\textbf{VoxelMorph~\cite{Balakrishnan}}} & \multicolumn{2}{|c}{\textbf{Baseline(8-1)}} & \multicolumn{2}{|c}{\textbf{FDRN(16-2)}}   \\ \midrule	
		-- & -- & -- & -- & w/o ADS & w/ ADS & w/o SL&w/ SL &w/o SL&w/ SL \\ \midrule	
		Dice &0.6079 &0.6805 &0.6689 &0.6746 &0.6897 &0.6764 &0.6898 &0.6882 &\textbf{0.7043} \\ \midrule
		NCC &0.9506 &0.9876 & 0.9962 &0.9973 &0.9971 &0.9969 &0.9966 &\textbf{0.9978} &0.9975\\ \midrule
		GPU/CPU(s) &--/-- &--/5658.19 &0.37/24.06 &\multicolumn{2}{|c}{0.26/18.02} &\multicolumn{2}{|c}{\textbf{0.14/7.93}} &\multicolumn{2}{|c}{0.29/25.62}\\
		\bottomrule 
	\end{tabular}
	\begin{tablenotes}
    \footnotesize
    SyN was conducted by ANTsPy and executed on the Intel(R) Xeon(R) E5-2650 v2 CPU.\\
    \end{tablenotes}
\end{threeparttable}
\end{table*}

\begin{figure*}[h]
	\centering
	\includegraphics[scale=0.35]{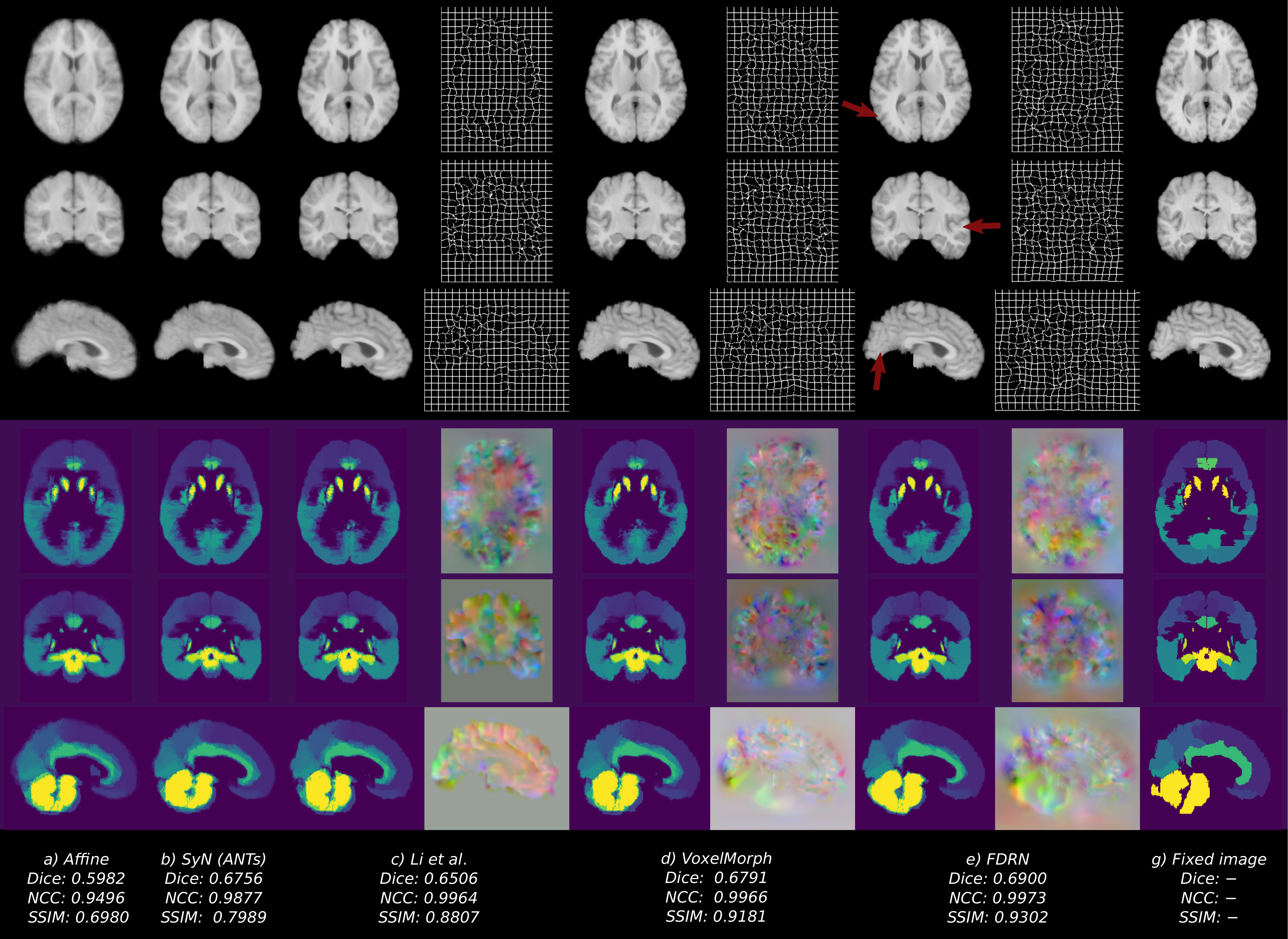}		
	\caption{Visual evaluation of different registration methods on LPBA40 MRI dataset. One of the 6 testing images was chosen as the fixed image and the rest 5 images were registered to it. The average of the 5 registered images and the corresponding labels are illustrated. Additionally, we exhibit the deformation field and the corresponding DVF of an image pair for demonstration purpose where the three dimensional DVF is represented in RGB channels.}
	\label{fig:LPBA40_View}
\end{figure*}

In order to evaluate FDRN visually, we selected one image from the 6 testing images in LPBA40 as the fixed image and registered the remaining 5 images to it. We illustrate the average of the 5 registered images along with the transformed labels with the mean Dice score, NCC and SSIM of different methods in Fig.~\ref{fig:LPBA40_View}. In addition, to visualize the individual registration performance, we demonstrate the deformation field and the corresponding DVF of an image pair for Li's method, VoxelMorph, and FDRN. It is shown that the average images of affine and SyN are severely blurred which indicates an inaccurate registration and a weak robustness against different variants of the moving image. Comparing to VoxelMorph, FDRN provides a sharper average image and more reliable registered labels which resemble the labels of the fixed image better.

\subsection{Evaluation on Unseen Brain MRI Datasets}\label{AA}
We evaluated the generalizability of the pretrained FDRN on different unseen brain MRI datasets including CUMC12, MGH10, ABIDE and ADNI. Particularly, we conducted standard preprocessing as mentioned in Section~\ref{sec:Dataset} and performed subject-to-subject registration for all the datasets where each of the images behaved as the fixed image and the remaining ones were registered to it. Experimental results are quantitatively summarized in Table~\ref{tab:unseen}. It is shown that our FDRN performs best in Dice and NCC in all the unseen datasets by resorting to the large network capacity and efficient learning.

\begin{table*}
	\centering
\begin{threeparttable}[b]
	\setlength{\tabcolsep}{1pt}
	\caption{Comparison of different deformable registration methods on the unseen MGH10, CUMC12, ABIDE and ADNI MRI datasets. Average Dice score of CUMC12 and MGH10 were calculated based on 7 segmented structures. 10 randomly chosen samples individually from ABIDE and ADNI were used for evaluation. Best results are in bold. (ADS was used in VoxelMorph.)}
	\label{tab:unseen}
	\begin{tabular}{c | c | c | c | c | c | c }
		\toprule
		\textbf{Dataset} & \textbf{Metrics} & \textbf{Affine} & \textbf{SyN} & \textbf{Li \textit{et al.}} & \textbf{VoxelMorph} & \textbf{FDRN}  \\ \midrule
 		MGH10&Dice/NCC &0.6474/0.7176 &0.6699/0.8615 &0.6726/0.8918 & 0.6678/0.9049 &\textbf{0.6865/0.9075}\\
		CUMC12&Dice/NCC&0.6111/0.6603 &0.6531/0.7931 &0.6617/0.8300 & 0.6517/0.8486 &\textbf{0.6669/0.8603} \\
		ABIDE &NCC &0.7793 &0.8739 & 0.9011 &0.9203 &\textbf{0.9324}  \\
   		ADNI &NCC &0.8078 &0.8852 & 0.9127 & 0.9279  &\textbf{0.9399} \\	
		\bottomrule 
	\end{tabular}
\end{threeparttable}
\end{table*}

\subsection{Model Analysis}\label{sec:MA}
\subsubsection{Model Variants}
\begin{figure*}[h]
	\centering
	 \includegraphics[scale=0.85]{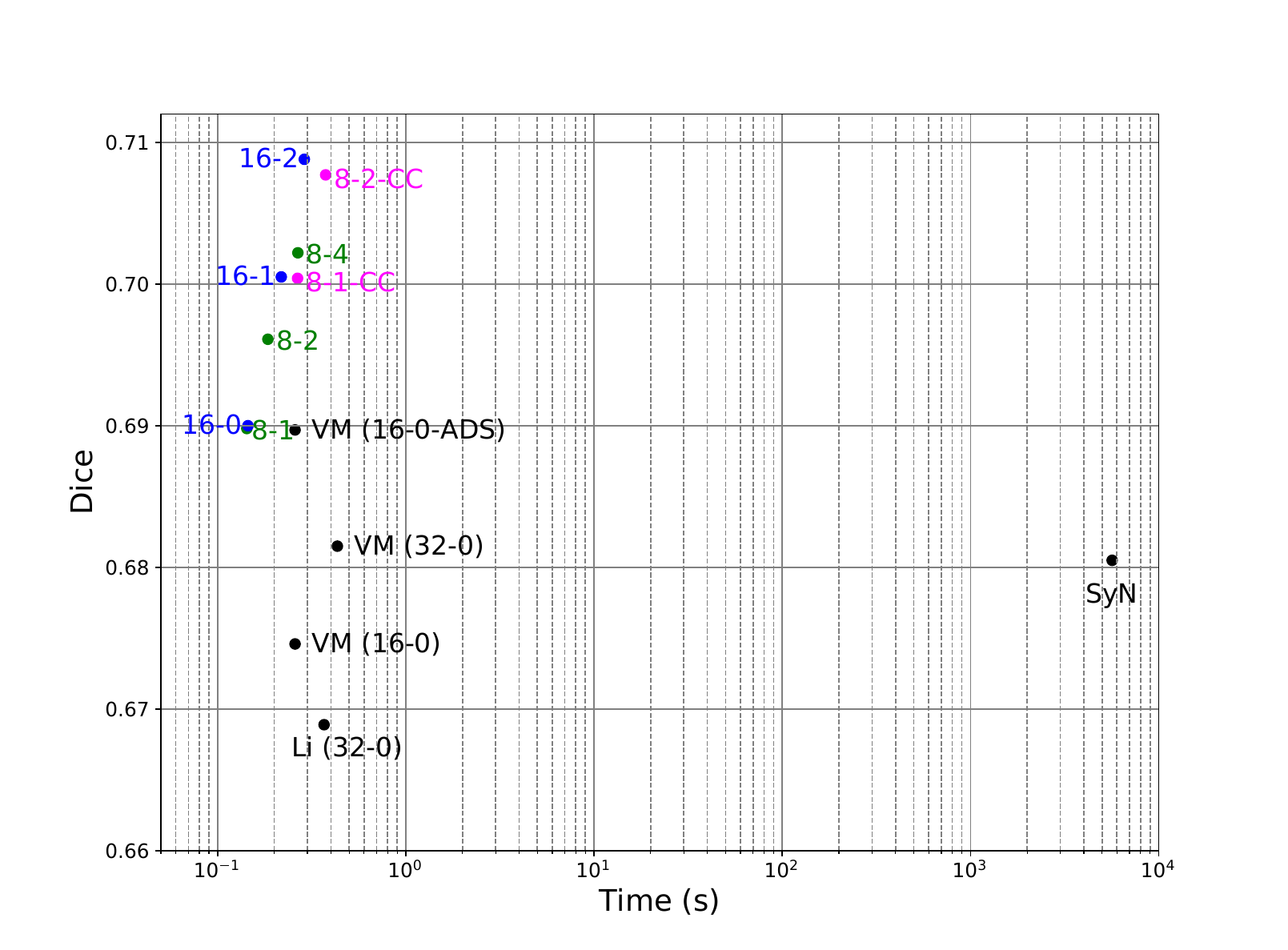}	
	\caption{Dice performance and runtime of different model variants on LPBA40 dataset with $\alpha_2=0.3$. (8-1: Baseline, 16-2: FDRN)}
	\label{fig:MV}
\end{figure*}

In order to analyze the capacity of the proposed network structure, we conducted experiments on different model variants in terms of the depth and width of the network. As illustrated in Fig.~\ref{fig:MV}, the green, blue, and magenta markers represent different variants of the proposed network architecture. Following the same notation manner in Section~\ref{sec:methods}, we denote the VoxelMorph in Table~\ref{tab:LPBA40} without and with ADS respectively as VM (16-0) and VM (16-0-ADS). The one with double features is indicated as VM(32-0) and Li's method is represented by Li (32-0). As expected, deepening and widening the network increase the model capacity and improve the Dice score. It is shown that deepening the model from 8-1 to 8-2 improves the Dice less than widening the channel to 16-1 because 16-1 has 2.4 times parameters as 8-2. Additionally, we show the performance of using channel concatenation (CC) instead of additive forwarding. We can observe that comparing to 8-1, 8-1 (CC) indeed improves Dice but it nearly doubles the training memory (about 10.2GB) with which we could almost adopt 16-2 (about 10.8GB), while 8-2 (CC) uses more than 13GB. Comparing to VM (16-0-ADS), our baseline model 8-1 achieves comparable Dice with nearly half of the runtime. 16-2 consumes similar runtime and training memory as VM (16-0-ADS) but contains 7.1 times model parameters. In Table~\ref{table:Param}, we summarize the amount of parameters required in different model variants.

\begin{table*}
\centering
\begin{threeparttable}[h]
\setlength{\tabcolsep}{2pt}
\caption{Number of required parameters in different networks. VM: VoxelMorph; CC: Channel concatenation. (8-1: Baseline, 16-2: FDRN)}
 \label{table:Param}
\begin{tabular}{c c c c c c c c c c c c }
    \toprule
    Models & 8-1 & 8-1(CC)& 8-2 & 8-2(CC)&8-4&16-0&16-1&16-2& Li(32-0)&VM(16-0)&VM(32-0)\\ \hline
	Param.& 285K&665K&466K& 1.0M& 830K& 397K &1.1M & 1.8M& 695K &252K &1.0M\\ 	
	\bottomrule 
   \end{tabular}
    \end{threeparttable}
\end{table*}

\subsubsection{Deep Supervision}
\begin{figure*}[h]
	\centering
	 \includegraphics[scale=0.9]{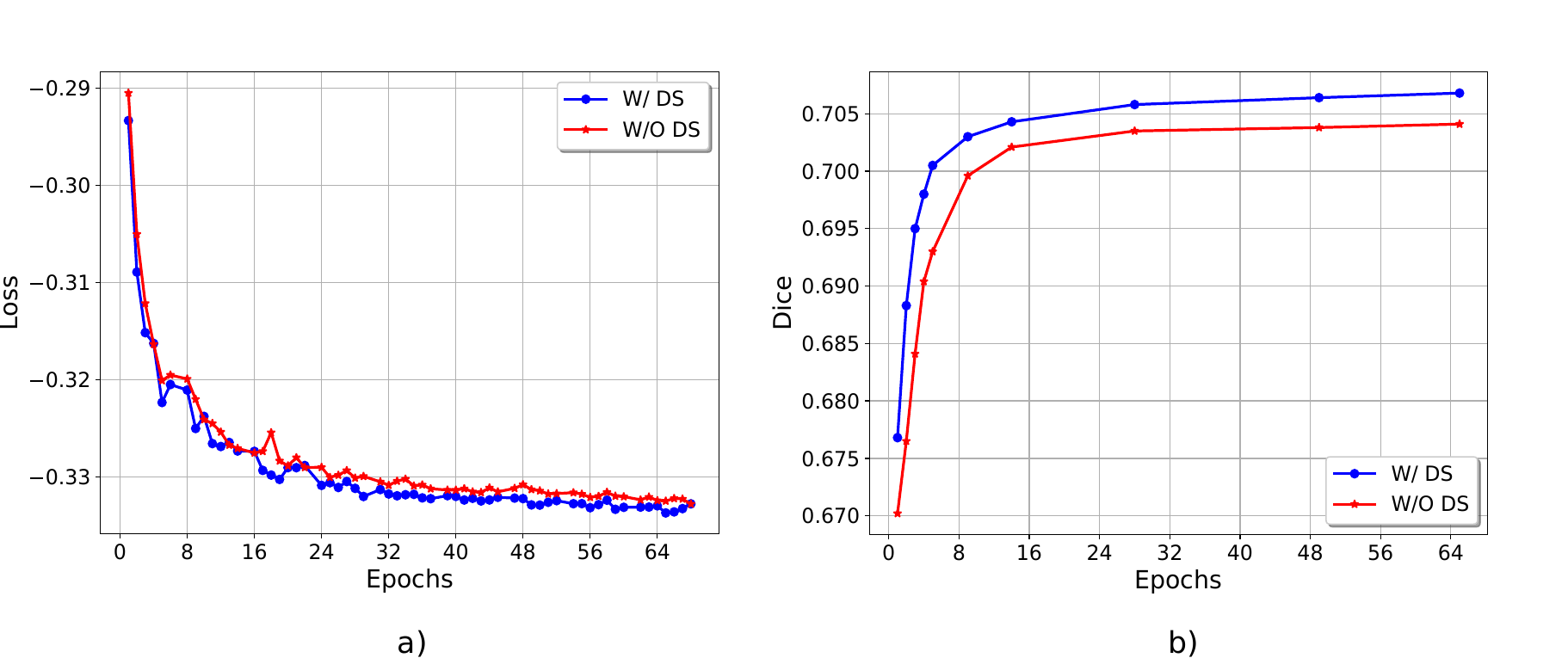}	
	\caption{Impact of deep supervision on the model convergence ($\alpha_2=0.3$): a) Loss function over epochs; b) Dice score over epochs.}
	\label{fig:DS}
\end{figure*}

In this section, we demonstrate the effectiveness of deep supervision on the convergence of FDRN. Particularly, we compare the proposed FDRN with the model variant without the LR loss. As mentioned in Section~\ref{sec:methods}, the LR loss is weighted by an exponentially decayed weighting factor $\lambda$ such that FDRN learns a rough registration by resorting to the LR images efficiently in the beginning and subsequently improves the registration accuracy fully based on the HR loss. In Fig.~\ref{fig:DS} a), we can see that the convergence rate of FDRN is noticeable faster than the variant without deep supervision since the early training phase. In the right figure, we demonstrate the impact of the LR loss on the Dice score over epochs. Comparing to adopting larger learning rate, the proposed deep supervision accelerates the convergence and meanwhile improves the registration accuracy. 

\subsubsection{Segmentation Loss}\label{sec:SL}
\begin{figure*}[h]
	\centering
	 \includegraphics[scale=0.9]{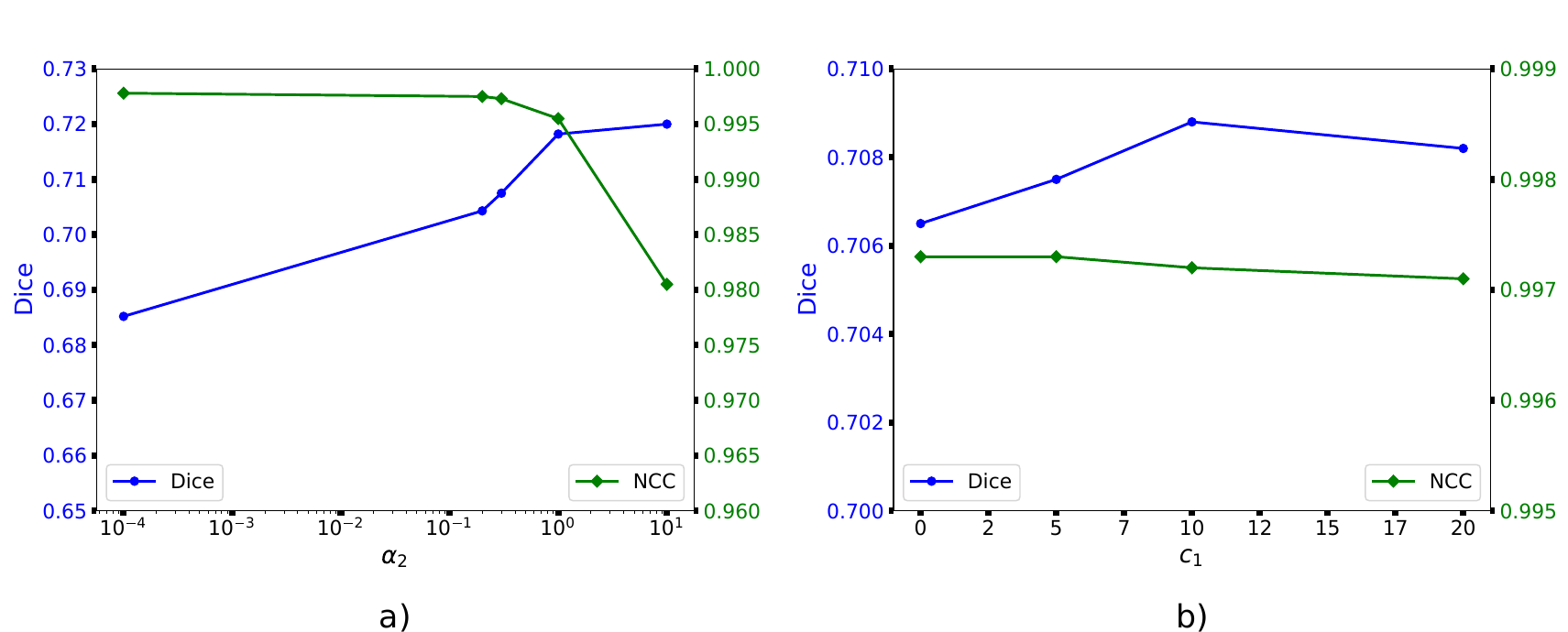}	
	\caption{Effectiveness of the segmentation loss: a) Weight $\alpha_2$ of the segmentation loss ($c_1=5$); b) Parameter $c_1$ of the segmentation loss ($\alpha_2=0.3$).}
	\label{fig:SL}
\end{figure*}

In order to analyze the proposed segmentation loss, we carried out experiments to evaluate the impact of the weighting factor $\alpha_2$ and the parameter $c_1$ in Eq.\eqref{Loss} on the Dice and NCC. As depicted in Fig.~\ref{fig:SL} a), when $\alpha_2$ increases, the segmentation loss gradually dominates the loss function which leads to a high Dice score. However, NCC drops severely when $\alpha_2>1$ which indicates that the segmentation loss might have overfitted the Dice score and result in an unrealistic DVF. In the experiment, we chose $\alpha_2 = 0.2$ for good performance of both Dice score and NCC. In the right figure, we can see that the Dice is fine tuned by $c_1$ and NCC seems rarely effected.

\subsubsection{Ablation Study}\label{sec:ablation}
In the ablation study, we analyze the behavior of the network by removing the proposed components including the additive forwarding (AF) linking the encoder path to the decoder counterpart, the residual learning (RL) within the encoder and decoder stages, the LR loss for deep supervision (DS), and the segmentation loss SL. The experiments were performed on the LPBA40 dataset and evaluated by average Dice score and NCC as depicted in Table~\ref {tab:ablation}. We can observe that AF improves both the Dice score and NCC by directly forwarding the extracted fine features. RL seems to contribute less to the Dice score and NCC than AF but it alleviates the gradient vanishing during the convergence. DS mainly accelerates the convergence rate and has a strong impact on the Dice score especially in the early training epochs. The proposed SL improves the Dice score by 1.93\%. In Section~\ref{sec:SL}, we have observed that by tunning the weight $\alpha_2$ and the parameter $c_1$ of SL, SL can achieve higher Dice score but may compromise the NCC. 
\begin{table}
	\centering
	\setlength{\tabcolsep}{4pt}
	\caption{Ablation study of the proposed FDRN based on average Dice score and NCC over all the segmented structures on the LPBA40 dataset with model 16-2 ($\alpha_2 = 0.3, c_1=5$). AF: Additive forwarding; RL: Residual learning; DS: Deep supervision; SL: Segmentation loss.}
	\label{tab:ablation}
	\begin{threeparttable}
	\begin{tabular}{c c c c c c c}
		\toprule
		AF  &\xmark &\cmark &\cmark &\cmark &\cmark \\ \midrule
		RL  &\cmark &\xmark &\cmark &\cmark &\cmark \\ \midrule
		DS  &\cmark &\cmark &\xmark &\cmark &\cmark \\ \midrule
		SL &\cmark &\cmark &\cmark &\xmark &\cmark \\ \midrule
		Dice/NCC &0.7001/0.9955 &0.7046/0.9972 &0.7048/0.9974 &0.6882/0.9978 &0.7075/0.9973\\  \bottomrule
	\end{tabular}
	\end{threeparttable}
\end{table}

%\clearpage
\section{DISCUSSION AND CONCLUSION}\label{sec:disconc}
The huge memory demand for 3D medical images limits the capacity of the registration network. In order to more efficiently exploit the memory resource, we propose a compact deformable registration network FDRN based on the autoencoder backbone which achieves better performance in both registration accuracy and runtime comparing to the investigated state-of-the-art methods including symmetric image normalization (SyN), Li's method, and VoxelMorph for brain MR images. Specially, our baseline model achieves comparable Dice as VoxelMorph and consumes nearly half of the runtime. FDRN further improves the registration accuracy by enlarging the model capacity and obtains a performance gain of 1.46\% in Dice comparing to VoxelMorph. Experiments show that the average of the registered images by FDRN obtains sharper anatomical structures than the other methods and the average transformed labels resemble the labels in the fixed image most which indicate that FDRN has a better registration accuracy and strong robustness against different variants of the moving image.

In order to explore the generalizability of the pretrained FDRN, we performed registration on the labeled CUMC12 and MGH10 MRI datasets and the unlabeled ABIDE and ADNI datasets. It is shown that FDRN obtains a large performance gain in Dice and NCC comparing to VoxelMorph in all the datasets. By resorting to the large network capacity and efficient learning, FDRN provides robust registration and achieves promising generalizability for the unseen images of inherent variability.

In the model analysis, we investigated the network architecture and demonstrate the contribution of different components of the model to the performance improvement. Specially, we have seen that increasing the model parameters improves the registration performance and model generalizability. Comparing to channel concatenation, double channel with additive forwarding has two major merits. First, it reduces the inference time without compromising the registration accuracy. Second, it contains more network parameters and saves more training memory which can be utilized for further enlarging the network capacity. Furthermore, we analyzed the impact of the low-resolution loss on the convergence curve. By leveraging an exponentially decreased weight during the training phase, the lower resolution image is involved to guide the learning in the beginning and subsequently the high-resolution loss dominates to fine tune the model. The proposed multi-label segmentation loss (SL) performs efficiently without inducing extra memory. Comparing to the memory-consuming one-hot encoding of multiple labels in average Dice score, the proposed SL can be employed on images with arbitrary amount of categories without compromising the depth of the network. In addition, we have shown that overusing the segmentation prior may cause degradation of NCC and result in unrealistic DVF. 

With regard to the computation time, the learning-based methods accomplish deformable registration of images of size 160$\times$208$\times$176 within 0.5s on the GPU and perform hundreds times faster than the traditional SyN on the CPU. Comparing to VoxelMorph, our baseline model halves the inference time and achieves comparable Dice. FDRN consumes similar runtime as VoxelMorph and contains 7.1 times parameters.

To conclude, the learning-based methods consume much less computation time than the traditional method SyN for a comparable registration accuracy. Experimental results demonstrate that the proposed compact model FDRN performs better than the state-of-the-art approach VoxelMorph on brain MRI datasets and shows good robustness to unseen images. It is necessary to mention that FDRN is a generalized model for deformable registration and is not limited to brain MR images. It can also be applied to other anatomical structures or CT images such as cardiac MR scans or lung CT images.

\section*{ACKNOWLEDGMENT}
This work has been funded by the Federal Ministry of Education and Research (BMBF, Germany) under the grand No. 05M18VSA.

\section*{CONFLICTS OF INTERES}
The authors have no conflict to disclose.

\section*{AVAILABILITY OF DATA}
The data that support the findings of this study are
available from the corresponding author upon reasonable
request.

\section*{REFERENCES}
\addcontentsline{toc}{section}{\numberline{}References}

% Following assumes you are using bibtex. However, for submission to the
% journal you MUST explicitly INCLUDE THE REFERENCES IN THE TEX FILE. 
% In that case you need the following

% \begin{thebibliography}{10}
% insert the .bbl file generated by bibtex here
	%This will be a series of entries from your .bib file formatted
	%something like
	%\bibitem{Me09}
        %{I.~Meijsing, B.~W.~Raaymakers, A.~J.~E.~Raaijmakers \it et al.},
        %\newblock {Dosimetry for the MRI accelerator: the impact of a 
	%magnetic field on the response of a Farmer NE2571 ionization chamber},
        %\newblock Phys. Med. Biol. {\bf 54}, 2993 -- 3002 (2009).

% \end{thebibliography}

% The following is when using bibtex and picks up the example.bib file

%\bibliography{Explicit address of .bib file}
\bibliography{./egbib.bib}      %example.bib is on the same directory

\begin{thebibliography}{10}

\bibitem{survey2}
J.~V. Hajnal, D.~L.~G. Hill, and D.~J. Hawkes,
\newblock {\em Medical Image Registration},
\newblock CRC press, 2001.

\bibitem{survey3}
B.~Fischer and J.~Modersitzki,
\newblock Ill-posed medicine--An introduction to image registration,
\newblock Inverse Problems {\bf 24}, 034008 (2008).

\bibitem{survey4}
A.~Sotiras, C.~Davatzikos, and N.~Paragios,
\newblock Deformable Medical Image Registration: A Survey,
\newblock IEEE Trans. Med. Imag. {\bf 32}, 1153--1190 (2013).

\bibitem{survey5}
Y.~Fu, Y.~Lei, T.~Wang, W.~J. Curran, T.~Liu, and X.~Yang,
\newblock Deep learning in medical image registration: a review,
\newblock Phys. Med. Biol.  (2020).

\bibitem{Beg2}
M.~F. Beg, M.~I. Miller, A.~Trouv\'{E}, and L.~Younes,
\newblock Computing large deformation metric mappings via geodesic flows of
  diffeomorphisms,
\newblock Int. J. Comput. Vision {\bf 61}, 139--157 (2005).

\bibitem{MI}
P.~Viola and W.~M.~W. III,
\newblock Alignment by maximization of mutual information,
\newblock Int. J. Comput. Vision {\bf 24}, 137--154 (1997).

\bibitem{Avants}
B.~B. Avants, C.~L. Epstein, M.~Grossman, and J.~C. Gee,
\newblock Symmetric diffeomorphic image registration with cross-correlation:
  Evaluating automated labeling of elderly and neurodegenerative brain,
\newblock Med. Image Anal. {\bf 12}, 26--41 (2008).

\bibitem{Bajcsy}
R.~Bajcsy and S.~Kova\v{c}i\v{c},
\newblock Multiresolution elastic matching,
\newblock Comput. Vis., Graph., Image Process. {\bf 46}, 1--21 (1989).

\bibitem{Gee}
J.~C. Gee and R.~Bajcsy,
\newblock Elastic matching: Continuum mechanical and probabilistic analysis,
\newblock Brain Warp. , 183--197 (1999).

\bibitem{Davatzikos}
C.~Davatzikos,
\newblock Spatial transformation and registration of brain images using
  elastically deformable models,
\newblock Comput. Vis. Image Understand {\bf 66}, 207--222 (1997).

\bibitem{Rueckert}
D.~Rueckert, L.~I. Sonoda, C.~Hayes, D.~L.~G. Hill, M.~O. Leach, and D.~J.
  Hawkes,
\newblock Nonrigid Registration Using Free--Form Feformation: Application to
  Breast MR Images,
\newblock IEEE Trans. Med. Imag. {\bf 18}, 712--721 (1999).

\bibitem{Kybic}
J.~Kybic and M.~Unser,
\newblock Fast parametric elastic image registration,
\newblock IEEE Trans. Image Process. {\bf 12}, 1427--1442 (2003).

\bibitem{Sdika}
J.~Kybic and M.~Unser,
\newblock A fast nonrigid image registration with constraints on the Jacobian
  using large scale constrained optimization,
\newblock IEEE Trans. Med. Imag. {\bf 27}, 271--281 (2008).

\bibitem{Thirion}
J.~Thirion,
\newblock Image matching as a diffusion process: an analogy with maxwell’s
  demons,
\newblock Med. Image Anal. {\bf 2}, 243--260 (1998).

\bibitem{Beg1}
M.~F. Beg and A.~Khan,
\newblock Symmetric data attachment terms for large deformation image
  registration,
\newblock IEEE Trans. Med. Imag. {\bf 26}, 1179--1189 (2007).

\bibitem{Ashburner2}
J.~Ashburner and K.~J. Friston,
\newblock Diffeomorphic registration using geodesic shooting and Gauss-Newton
  optimisation,
\newblock NeuroImage {\bf 55}, 954--967 (2011).

\bibitem{rueckert1}
D.~Rueckert, A.~F. Frangi, and J.~A. Schnabel,
\newblock Automatic construction of 3D statistical deformation models using
  non-rigid registration,
\newblock in {\em Proc. Int. Conf. Med. Imag. Comp. Comput. Assist. Interv.
  (MICCAI)}, pages 77--84, 2001.

\bibitem{rueckert2}
D.~Rueckert, A.~F. Frangi, and J.~A. Schnabel,
\newblock Automatic construction of 3-D statistical deformation models of the
  brain using nonrigid registration,
\newblock IEEE Trans. Med. Imag. {\bf 22}, 1014--1025 (2003).

\bibitem{krebs}
J.~Krebs et~al.,
\newblock Robust non-rigid registration through agent-based action learning,
\newblock in {\em Proc. Int. Conf. Med. Imag. Comp. Comput. Assist. Interv.
  (MICCAI)}, pages 344--352, 2017.

\bibitem{gutierrez1}
B.~Guti{\'e}rrez-Becker, D.~Mateus, L.~Peter, and N.~Navab,
\newblock Learning optimization updates for multimodal registration,
\newblock in {\em Proc. Int. Conf. Med. Imag. Comp. Comput. Assist. Interv.
  (MICCAI)}, pages 19--27, 2016.

\bibitem{gutierrez2}
B.~Guti{\'e}rrez-Becker, D.~Mateus, L.~Peter, and N.~Navab,
\newblock Guiding multimodal registration with learned optimization updates,
\newblock Med. Image Anal. {\bf 41}, 2--17 (2017).

\bibitem{kim}
M.~Kim, G.~Wu, Q.~Wang, S.~Lee, and D.~Shen,
\newblock Improved image registration by sparse patch-based deformation
  estimation,
\newblock NeuroImage {\bf 105}, 257--268 (2015).

\bibitem{wang}
Q.~Wang, M.~Kim, Y.~Shi, G.~Wu, and D.~Shen,
\newblock Predict brain MR image registration via sparse learning of appearance
  and transformation,
\newblock Med. Image Anal. {\bf 20}, 61--75 (2015).

\bibitem{fan}
J.~Fan, X.~Cao, P.~Yap, and D.~Shen,
\newblock BIRNet: Brain image registration using dual-supervised fully
  convolutional networks,
\newblock Med. Image Anal. {\bf 54}, 193--206 (2019).

\bibitem{sokooti}
H.~Sokooti, B.~de~Vos, F.~Berendsen, B.~P.~F. Lelieveldt, I.~I\v{s}gum, and
  M.~Staring,
\newblock Nonrigid image registration using multi-scale 3D convolutional neural
  networks,
\newblock in {\em Proc. Int. Conf. Med. Imag. Comp. Comput. Assist. Interv.
  (MICCAI)}, pages 232--239, 2017.

\bibitem{yang}
X.~Yang, R.~Kwitt, M.~Styner, and M.~Niethammer,
\newblock Quicksilver: Fast predictive image registration--a deep learning
  approach,
\newblock NeuroImage {\bf 158}, 378--396 (2017).

\bibitem{rohe}
M.~M. Roh{\'e}, M.~Datar, T.~Heimann, M.~Sermesant, and X.~Pennec,
\newblock SVF-Net: Learning deformable image registration using shape matching,
\newblock in {\em Proc. Int. Conf. Med. Imag. Comp. Comput. Assist. Interv.
  (MICCAI)}, pages 266--274, 2017.

\bibitem{hu2018}
Y.~Hu et~al.,
\newblock Weakly-supervised convolutional neural networks for multimodal image
  registration,
\newblock Med. Image Anal. {\bf 49}, 1--13 (2018).

\bibitem{deVos}
B.~D. deVos, F.~F. Berendsen, M.~A. Viergever, M.~Staring, and I.~I\v{s}gum,
\newblock End-to-end unsupervised deformable image registration with a
  convolutional neural network,
\newblock in {\em Proc. Deep Learning Med. Imag. Anal. Multimodal Learning for
  Clinical Decision Support}, pages 204--212, 2017.

\bibitem{Balakrishnan}
G.~Balakrishnan, A.~Zhao, M.~R. Sabuncu, J.~Guttag, and A.~V. Dalca,
\newblock An unsupervised learning model for deformable medical image
  registration,
\newblock in {\em Proc. IEEE Conf. Comput. Vis. Pattern Recognit.}, pages
  9252--9260, 2018.

\bibitem{Li}
H.~Li and Y.~Fan,
\newblock Non-Rigid Image Registration Using Self-Supervised Fully
  Convolutional Networks without Training Data,
\newblock in {\em Proc. IEEE Int. Symp. Biomed. Imaging}, pages 1--4, 2018.

\bibitem{Jaderberg}
M.~Jaderberg, K.~Simonyan, and A.~Zisserman,
\newblock Spatial transformer networks,
\newblock in {\em Adv. Neural. Inf. Process. Syst.}, pages 2017--2025, 2015.

\bibitem{unet}
O.~Ronneberger, P.~Fischer, and T.~Brox,
\newblock U-net: Convolutional networks for biomedical image segmentation,
\newblock in {\em Proc. Int. Conf. Med. Imag. Comp. Comput. Assist. Interv.
  (MICCAI)}, pages 234--241, Springer, 2015.

\bibitem{V2}
G.~Balakrishnan, A.~Zhao, M.~R. Sabuncu, J.~Guttag, and A.~V. Dalca,
\newblock Voxelmorph: a learning framework for deformable medical image
  registration,
\newblock IEEE Trans. Med. Imag. {\bf 38}, 1788--1800 (2019).

\bibitem{LPBA40}
D.~W. Shattuck et~al.,
\newblock Construction of a 3D probabilistic atlas of human cortical
  structures,
\newblock Neuroimage {\bf 39}, 1064--1080 (2008).

\bibitem{klein}
A.~Klein et~al.,
\newblock Evaluation of 14 nonlinear deformation algorithms applied to human
  brain {MRI} registration,
\newblock Neuroimage {\bf 46}, 786--802 (2009).

\bibitem{ABIDE}
A.~D. Martino et~al.,
\newblock The autism brain imaging data exchange: towards a large-scale
  evaluation of the intrinsic brain architecture in autism,
\newblock Molecular psychiatry {\bf 19}, 659--667 (2014).

\bibitem{ADNI}
S.~G. Mueller et~al.,
\newblock Ways toward an early diagnosis in Alzheimer’s disease: the
  Alzheimer’s Disease Neuroimaging Initiative (ADNI),
\newblock Alzheimer's \& Dementia {\bf 1}, 55--66 (2005).

\bibitem{unbiased}
V.~Fonov et~al.,
\newblock Unbiased average age-appropriate atlases for pediatric studies,
\newblock Neuroimage {\bf 54}, 313--327 (2011).

\bibitem{Dice}
L.~R. Dice,
\newblock Measures of the amount of ecologic association between species,
\newblock Ecology {\bf 26}, 297--302 (1945).

\bibitem{ssim}
Z.~Wang, A.~C. Bovik, H.~R. Sheikh, and E.~P. Simoncelli,
\newblock Image quality assessment: from error visibility to structural
  similarity,
\newblock IEEE Trans. Image Process. {\bf 13}, 600--612 (2004).

\bibitem{ANTs}
B.~B. Avants, N.~J. Tustison, G.~Song, P.~A. Cook, A.~Klein, and J.~C. Gee,
\newblock A reproducible evaluation of ANTs similarity metric performance in
  brain image registration,
\newblock Neuroimage {\bf 54}, 2033--2044 (2011).

\end{thebibliography}
% above points to where we find the master reference list
% and also causes the bibliography to be printed

% When creating your bibliography you should run bibtex on your local
% computer after running pdflatex on your .tex file. bibtex will
% generate a .bbl file.
% Copy the contents of this .bbl file into your main latex document,
% replacing the "\bibliography" command which was pointing at your .bib file.

% following defines style of .bbl file 

%\bibliographystyle{explicit relative path to medphy.bst}
\bibliographystyle{./medphy.bst}    %if this is installed on your system,
				    %it is not essential to have the    ./

% Note that you need to typeset once, then run bibtex, then typeset another
% two times to get the references working properly.
% \end{multicols}

\end{document}